# Translating Expert Intuition into Quantifiable Features: Encode Investigator Domain Knowledge via LLM for Enhanced Predictive Analytics

Phoebe Jing, Yijing Gao, Yuanhang Zhang, Xianlong Zeng[1]

## Abstract

In the realm of predictive analytics, the nuanced domain knowledge of investigators often remains underutilized, confined largely to subjective interpretations and ad hoc decision-making. This paper explores the potential of Large Language Models (LLMs) to bridge this gap by systematically converting investigator-derived insights into quantifiable, actionable features that enhance model performance. We present a framework that leverages LLMs' natural language understanding capabilities to encode these red flags into a structured feature set that can be readily integrated into existing predictive models. Through a series of case studies, we demonstrate how this approach not only preserves the critical human expertise within the investigative process but also scales the impact of this knowledge across various prediction tasks. The results indicate significant improvements in risk assessment and decision-making accuracy, highlighting the value of blending human experiential knowledge with advanced machine learning techniques. This study paves the way for more sophisticated, knowledge-driven analytics in fields where expert insight is paramount.

## Introduction

In the rapidly evolving domain of predictive analytics, combining human intuition with advanced computational technologies offers a promising frontier for research and application. Traditional predictive models primarily rely on quantitative data; however, the complex nature of human decision-making, particularly in specialized fields such as financial fraud detection [2,11], healthcare [3-10], and national security, often hinges on qualitative insights derived from years of expert experience. This nuanced understanding, referred to as 'expert intuition', encompasses an investigator's ability to detect subtle, non-obvious signals that may not be readily apparent in the data but are crucial for accurate predictions and decision-making.

The inherent value of this expert intuition lies in its deep, context-specific knowledge, which can significantly enhance the effectiveness of predictive models. Despite its importance, the integration of such qualitative insights into quantitative frameworks remains a significant challenge in analytics. Current methodologies struggle to capture the abstract and often tacit knowledge that experts possess, primarily because this information is subjective and difficult to standardize and quantify.

[1] xz926813@ohio.edu

Recent advancements in data science have made substantial strides in addressing these challenges, with machine learning models increasingly able to process and analyze large volumes of unstructured data [1]. However, these technologies often fail to capture the depth of human expertise, as the subtle nuances of expert intuition are not easily translated into the rigid structures required by traditional algorithms. As a result, much of the domain-specific knowledge remains underutilized, confined to ad-hoc decision-making and subjective interpretation.

Moreover, the academic literature reveals a gap in the systematic conversion of expert insights into actionable, measurable features within predictive models. While there are methodologies for incorporating expert knowledge, such as expert systems and rule-based algorithms, these often do not leverage the latest advancements in natural language processing and machine learning, thereby limiting their effectiveness and scalability.

This paper proposes a novel framework that addresses these challenges by utilizing Large Language Models (LLMs) to encode expert-derived insights into structured, actionable features. LLMs, with their advanced natural language understanding capabilities, offer a unique opportunity to bridge the gap between qualitative expert knowledge and quantitative predictive models. By transforming qualitative 'red flags' into quantifiable metrics, our approach allows for the seamless integration of expert insights into existing predictive analytics frameworks.

The application of LLMs in this context is twofold. Firstly, LLMs can analyze and interpret the natural language in which expert knowledge is often articulated, thereby capturing the subtleties and complexities of this information. Secondly, through machine learning techniques such as feature extraction and sentiment analysis, LLMs can transform these interpretations into standardized formats that predictive models can utilize effectively.

The primary contribution of this paper is the development and presentation of an innovative framework that leverages Large Language Models (LLMs) to integrate expert intuition into predictive analytics. This framework systematically encodes qualitative expert insights, commonly referred to as 'red flags', into quantifiable features that can be utilized by predictive models. By employing the advanced natural language processing capabilities of LLMs, we offer a methodological advancement in how qualitative data can be standardized and operationalized in a quantitative analytic environment. Our key contributions are shown below:
- **Proposing a Novel Framework**: We introduce a new approach that utilizes the natural language understanding capabilities of LLMs to capture and encode the nuanced, domain-specific knowledge of experts into actionable, structured data formats. This framework provides a structured pathway for transforming qualitative insights into measurable variables that predictive models can incorporate.
- **Demonstration through Examples**: To illustrate the potential of our proposed framework, we present a series of conceptual examples that highlight how expert insights can be transformed into structured features. These examples are designed to showcase the framework's applicability and versatility across different domains without conducting rigorous empirical testing.
- **Highlighting Future Research Directions**: While this paper does not empirically validate the predictive power enhancement of the models through the proposed methodology, it sets the groundwork for future studies. We outline specific areas where subsequent research can apply empirical methods to validate and refine the framework, such as testing in various industry settings, comparing the performance of different LLM configurations, and exploring ethical implications.

# Method

This study introduces a novel methodology utilizing Large Language Models (LLMs) to systematically transform qualitative insights from financial investigators into quantifiable, actionable features that can enhance predictive analytics models. The focus is on leveraging the natural language understanding capabilities of LLMs to identify and encode patterns indicative of complex activities such as money laundering. Our approach involves constructing detailed prompts based on transaction data, which guide the LLM in extracting relevant features from these data points. Figure 1 illustrates the methodology of our approach for encoding expert insights into predictive models using LLMs.

## I. Data Collection and Preparation

The initial phase of our methodology involves the collection and preparation of transaction data, which typically includes details such as transaction dates, account numbers, transaction amounts, currencies, and transaction types. For this study, we specifically look at patterns that suggest 'fan-out' money laundering activities, where large sums are rapidly dispersed across multiple accounts. Investigators provide these datasets, highlighting transactions suspected of embodying this pattern, thus setting the stage for the subsequent analysis.

## II. Prompt Construction and Feature Identification

With the data prepared, the next step is to construct prompts that encapsulate the specific laundering patterns under investigation. These prompts are carefully designed to highlight the nuances of the transactions that may indicate suspicious activities. For example, a prompt may describe the rapid movement of funds from a single source to multiple destinations within a short timeframe, a common tactic in money laundering. The LLM uses these prompts to focus its analysis, applying its trained natural language processing capabilities to dissect and understand the complex descriptions within the transaction data.

## III. LLM Processing and Feature Extraction

Upon receiving the prompt and associated transaction data, the LLM processes this information to identify and extract key features indicative of money laundering. The model analyzes the data to determine features such as the number of linked transactions, the diversity of the transaction amounts, the variety of currencies used, and the time intervals between transactions. This step is crucial as it transforms qualitative assessments into quantitative features that can be readily incorporated into predictive models.

### IV. Quantification and Integration into Predictive Models

The final step in our methodology involves quantifying the insights extracted by the LLM. Each identified feature is assigned a numerical value that reflects its significance in the context of predictive analytics. These quantified features are then integrated into existing predictive models to test their efficacy in enhancing the predictive accuracy of these models. The integration process is designed to be seamless, ensuring that the enriched models can utilize the new features without requiring substantial modifications to their existing structures.

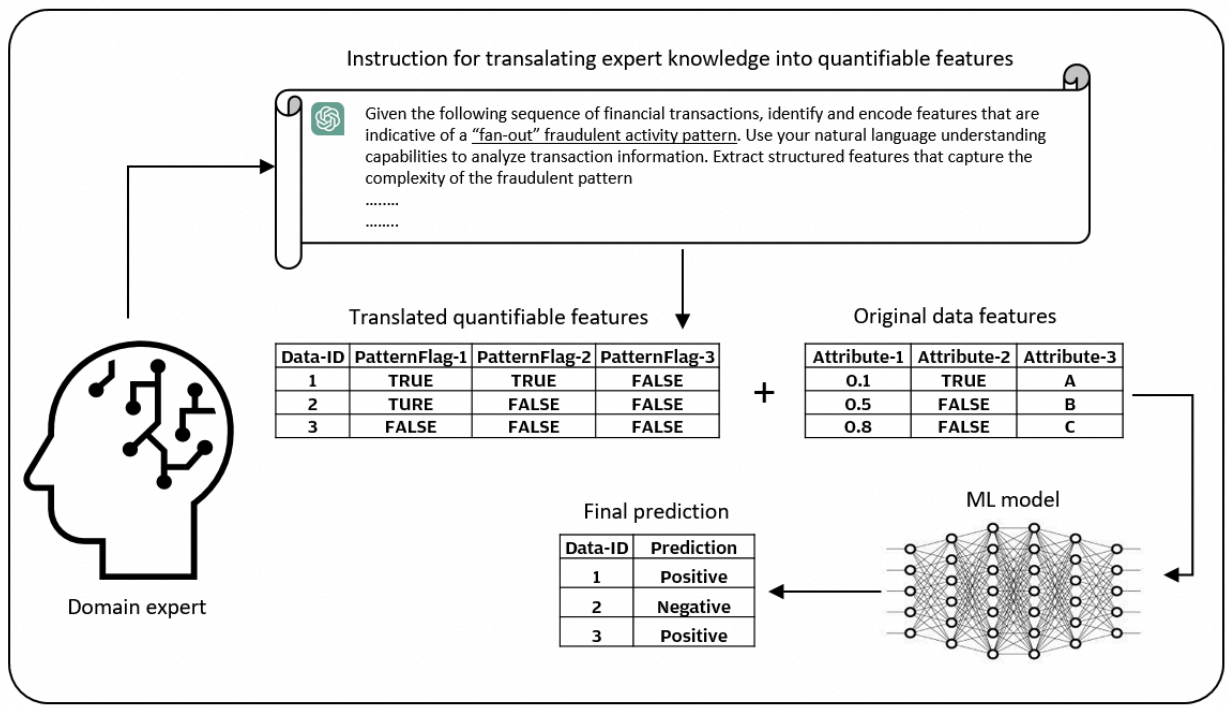

Figure 1. A high level illustrates the methodology of our approach for encoding expert insights into predictive models using LLMs.

# Limitation and Future Work

## Limitation

This study, while pioneering in its approach to integrating Large Language Models (LLMs) with expert intuition for predictive analytics, is subject to several limitations that must be acknowledged:
- **Lack of Empirical Validation**: The primary limitation of this study is the absence of empirical testing of the proposed framework. As the current research focuses on conceptual development and illustrative examples, the actual enhancement in predictive power through the integration of encoded expert insights remains theoretical. Without quantitative testing, the effectiveness of the framework in improving decision-making accuracy and model performance is not demonstrated.
- **Scope of Example Applications**: The examples presented in this study are illustrative and are designed to demonstrate the framework's conceptual viability rather than its effectiveness across real-world scenarios. These examples may not capture the complex variability and challenges
encountered in actual predictive tasks across different industries.
- **Complexity of Expert Knowledge Encoding**: The process of encoding expert knowledge into quantifiable features involves assumptions and simplifications that might not fully capture the depth and subtlety of the original insights. There is a risk that crucial nuances of expert knowledge could be lost or misrepresented during this translation process. Dependency on Language Model Capabilities: The framework's effectiveness is heavily dependent on the capabilities of the underlying LLMs. Limitations inherent to these models, such as biases in training data or errors in natural language understanding, could adversely affect the quality and reliability of the feature sets generated.

## Future Work

To address these limitations and extend the research, the following future work is proposed:
- **Empirical Testing**: Future studies should focus on empirically testing the framework within various predictive analytics tasks across multiple industries. This would involve integrating the encoded features into predictive models and quantitatively assessing the changes in their performance. Such testing would provide valuable insights into the practical benefits and limitations of the framework.
- **Enhancement of Encoding Techniques**: Further research is needed to refine the methods used to encode expert knowledge into structured data. This could involve developing more sophisticated algorithms that can capture a greater level of detail and subtlety from the expert insights, potentially incorporating machine learning techniques that can learn and adapt from feedback.
- **Exploration of Ethical and Bias Considerations**: It is crucial to examine the ethical implications of automating expert knowledge, particularly concerning biases that may be present in the data or introduced by the models. Future work should include the development of methodologies to identify, mitigate, and monitor these biases to ensure fair and ethical use of the technology.
- **Integration of Multimodal Data**: Considering the integration of other forms of data, such as images or sensor data, alongside textual expert insights could enhance the richness of the feature sets and the accuracy of the predictive models. Research into multimodal learning approaches would be beneficial in this context.